\documentclass[letterpaper, 10 pt, conference]{ieeeconf}  

\usepackage{cite}
\usepackage{amsmath,amssymb,amsfonts}
\usepackage{algorithmic}
\usepackage{graphicx}
\usepackage{textcomp}
\usepackage{xcolor}
\usepackage{physics}
\usepackage{float}
\usepackage{bm}
\usepackage{bbm}
\usepackage{svg}
\usepackage{subcaption}
\usepackage{mathtools}

\IEEEoverridecommandlockouts                              

\title{\LARGE \bf
Privacy-Preserving Decentralized Cooperative Localization with Range-Only Measurements: A Convex Optimization Based Approach
}

\author{Nitesh Kumar$^{1}$, Reyshwanth Ganeshan$^{1}$, Sixu Li$^{2,*}$, Sivakumar Rathinam$^{1,3}$ and Swaroop Darbha$^{1}$
\thanks{$^{1}$Department of Mechanical Engineering, Texas A\&M University, College Station, TX 77843, USA
        {\tt\small  niteshk@tamu.edu; reyshwanth@tamu.edu; srathinam@tamu.edu; dswaroop@tamu.edu }}%
\thanks{$^{2}$Zachry Department of Civil \& Environmental Engineering, Texas A\&M University, College Station, TX 77843, USA
        {\tt\small  sixuli@tamu.edu}}%
\thanks{$^{3}$Computer Science \& Engineering, Texas A\&M University, College Station, TX 77843 USA}%
\thanks{$^*$Corresponding author: Sixu Li
}%
}

\begin{document}

\maketitle
\thispagestyle{empty}
\pagestyle{empty}

\begin{abstract}

Cooperative localization using range-based measurements is critical for multi-robot systems operating in GPS-denied and unstructured environments. However, traditional cooperative approaches require sharing explicit spatial coordinates across the network, presenting a severe security vulnerability in privacy-sensitive missions. While recent literature has explored privacy-preserving alternatives, these methods typically rely on accuracy-degrading noise injection or computationally prohibitive cryptographic protocols. To overcome these limitations, we propose a novel, natively privacy-preserving Decentralized Cooperative Localization (DCL) framework based on convex optimization. Discarding probabilistic noise models, we assume strictly bounded measurement noise and formulate the localization problem via Semi-Definite Programming (SDP) to compute a Maximum-Volume Inscribed Ellipsoid (MVE). Our approach introduces novel intersection-plane constraints derived from landmark measurements to significantly tighten individual spatial bounds. To incorporate inter-robot range measurements securely, we uniquely decompose coupling constraints into localized Linear Matrix Inequalities (LMIs). Agents achieve fleet-wide spatial consensus by iteratively exchanging only abstract dual variables, completely avoiding the transmission of explicit primal position estimates. Extensive 3D Monte Carlo simulations demonstrate that our DCL framework outperforms existing SDP-based localization method in accuracy, while guaranteeing operational privacy and maintaining highly scalable, parallelizable computation.
\end{abstract}

\section{Introduction}
\label{sec:introduction}

The deployment of multi-robot systems in unstructured, GPS-denied environments, such as underwater, indoor, and subterranean domains, requires robust spatial awareness for safe, autonomous operations. In standard robotic navigation architectures, high-frequency proprioceptive data—such as odometry and Inertial Measurement Unit (IMU) readings—are continuously fused with absolute position updates using recursive estimators like the Extended Kalman Filter (EKF)~\cite{moore2015generalized}. Just as GPS offers global reference points outdoors, range-based relative measurements can serve as an alternative source of absolute position correction in GPS-denied environments. Consequently, cooperative localization~\cite{roumeliotis2002distributed,fox2000probabilistic,lajoie2022towards} allows teams of robots to leverage ubiquitous range sensors—such as underwater acoustic modems or Ultra-Wideband (UWB) radios—to fuse local sensor data with inter-robot measurements, generating the robust, divergence-free position estimates essential for long-term navigational stability~\cite{gonzalez2020autonomous}.

However, as safety and security concerns become increasingly prominent~\cite{SARTAYEVA2023103293,7321972,denning2009spotlight,oruma2023security}, the issue of spatial privacy in these cooperative networks has emerged as a critical research frontier. Because traditional cooperative architectures rely on the continuous inter-robot communication of their estimated positions, they are inherently susceptible to information leakage~\cite{tang2025feasibility}. Broadcasting explicit spatial coordinates allows adversaries to track fleet trajectories, creating severe privacy risks for sensitive operations like defense, resource extraction, and emerging civilian Urban Air Mobility (UAM) networks.

\begin{figure}[t!]
\centering
\includegraphics[width=0.95\linewidth]{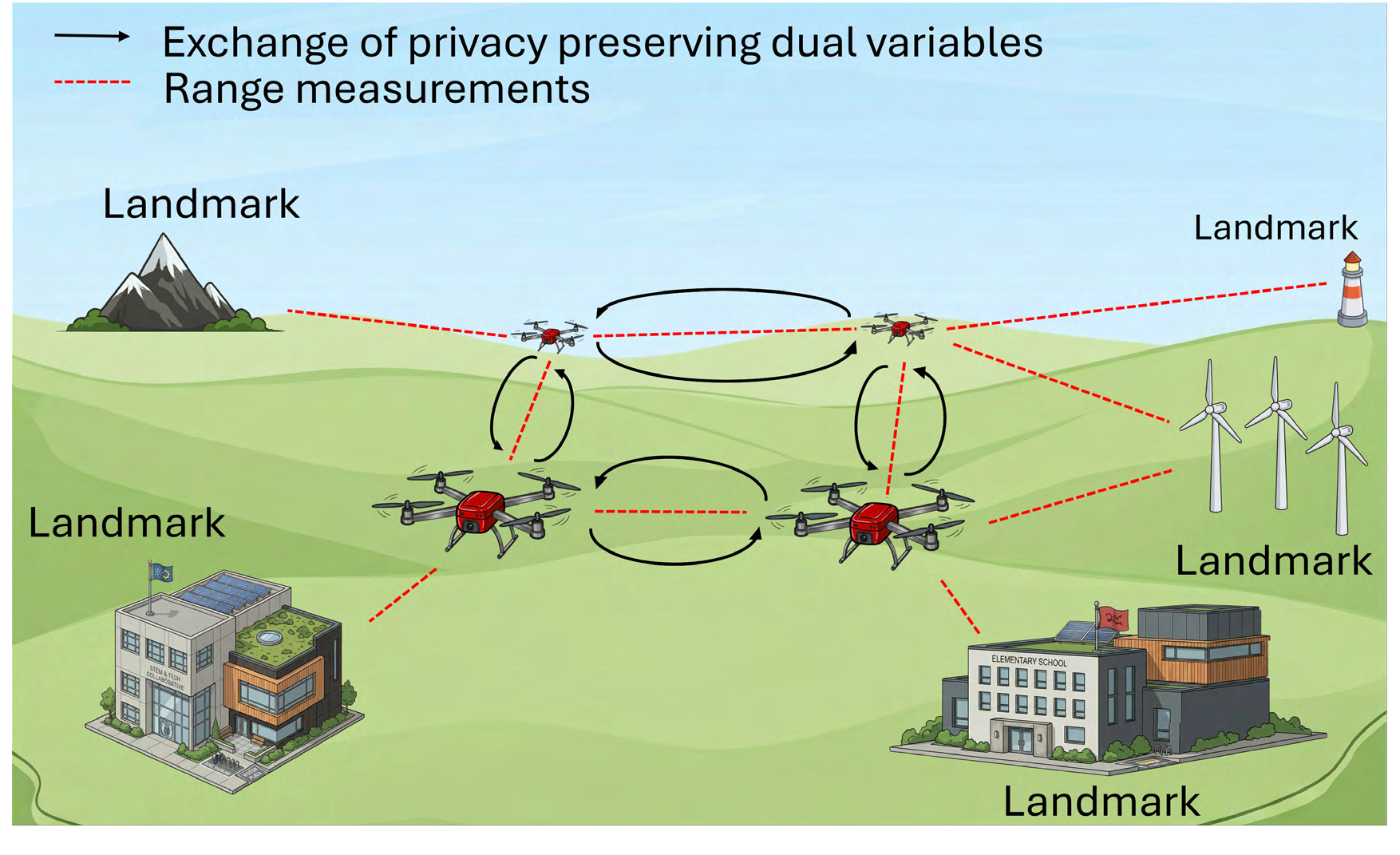}
\caption{Illustration of privacy-preserving decentralized cooperative localization: robot swarms cooperating using range-based measurements, exchanging only anonymized penalty signals (dual variables) rather than positions.}
\label{fig:privacy_scenario}
\end{figure}

Achieving high-accuracy, range-based cooperative localization \emph{without} sacrificing each agent's spatial privacy remains a significant, unresolved challenge. Consequently, \emph{privacy-preserving range-based cooperative localization} has emerged as a critical research area within robotics and sensor networks~\cite{yu2023balancing,shit2022privacy,le2025privacy,zhu2024protecting}. Existing algorithmic frameworks designed to address this challenge largely fall into two paradigms, each with notable limitations. Perturbation-based methods, such as geo-indistinguishability or differential privacy~\cite{yu2023balancing,shit2022privacy}, achieve privacy by injecting artificial noise into the shared data, which inherently degrades overall estimation accuracy. Conversely, cryptography-based techniques employing homomorphic encryption or secret sharing~\cite{le2025privacy,zhu2024protecting} preserve data fidelity but introduce severe computational and communication burdens that render them impractical for real-time, resource-constrained swarms.

Another major limitation is the reliance on specific noise models, typically assuming zero-mean Gaussian errors. Real-world phenomena, such as underwater acoustic ranging, routinely exhibit asymmetric and heavy-tailed noise due to multipath fading, non-line-of-sight (NLOS) conditions, or environmental effects~\cite{rs17152637}. In practice, it is often more tractable to bound the noise than to characterize its distribution~\cite{angelopoulos2022gentleintroductionconformalprediction,pmlr-v242-tang24a}. 

A {\it distinguishing departure of this work} from prior work is as follows: \textbf{We assume \emph{strict bounds} on the noise from range measurements. This mild and practical assumption enables us to construct deterministic, convex feasible sets, providing robustness against unpredictable anomalies.} By bounding each robot's feasible region with a Maximum-Volume Inscribed Ellipsoid (MVE), we explicitly quantify both the position and the localization uncertainty simultaneously~\cite{rs17152637,boyd2004convex}. Semi-Definite Programming (SDP) formulations yield \cite{rs17152637} robust localization in the presence of bounded noise, but traditional SDP approaches require a central solver that collects all robots' positions, creating both scalability bottlenecks and privacy risks. 

To overcome these barriers, we propose {\it a novel, privacy-preserving decentralized SDP based framework.} Our approach enables robots to achieve accurate, secure fleet-wide localization consensus by exchanging only local resource matrices and associated dual variables~\cite{boyd2011distributed}---mathematical penalty signals that cannot be inverted to reveal positions. \emph{No robot ever transmits its position estimate}, completely bypassing the need for privacy-degrading noise or heavy cryptography. In summary, our main contributions are:
\begin{enumerate}
    \item We introduce novel intersection-plane constraints, derived from  measurements to landmark pairs, that substantially tighten each agent's feasible region.
    \item We develop an iterative, decentralized SDP framework, in which only dual variables (not explicit positions) are exchanged, natively preserving agent privacy.
    \item We deliver a comprehensive simulation framework benchmarking our decentralized SDP against both a centralized oracle and independent (non-cooperative) baselines, demonstrating near-centralized accuracy and robust, native privacy protection.
\end{enumerate}

\section{Problem Formulation}

Consider a system of $N$ robots and $M$ static landmarks embedded in $\mathbb{R}^3$. Let $\mathbf{r}_i \in \mathbb{R}^3$ denote the (unknown) position of robot $i \in \{1,\dots,N\}$, and let $\mathbf{B}_k \in \mathbb{R}^3$ denote the (known) position of landmark $k \in \{1,\dots,M\}$. For each robot $i$, let $\mathcal{N}_i^{\text{A}} \subseteq \{1,\dots,M\}$ and $\mathcal{N}_i^{\text{R}} \subseteq \{1,\dots,N\}$ denote the respective sets of static landmarks and neighboring robots that fall within its sensing range.

\subsection{Robot–landmark Measurements}

When robot $i$ measures its range to landmark $j$, we model that the true distance lies within the measured distance interval:
\begin{equation}
    \underline{\rho}_{ij} \;\le\; \|\mathbf{r}_i - \mathbf{B}_j\| \;\le\; \bar{\rho}_{ij},
    \quad j \in \mathcal{N}_i^{\text{A}},
    \label{eq:robot-landmark-interval}
\end{equation}
where $\underline{\rho}_{ij}$ and $\bar{\rho}_{ij}$ represent the lower and upper bounds, respectively. The upper bounds $\bar{\rho}_{ij}$ generate convex outer-sphere constraints. Conversely, the lower bounds $\underline{\rho}_{ij}$ are utilized to construct planar constraints by leveraging the outer-sphere constraints from other landmarks, which effectively tightens the overall feasible set.

\subsection{Robot–Robot Measurements}

For a range measurement between robots $i$ and $\ell$, the true distance is less than the measured value 
\begin{equation}
    \|\mathbf{r}_i - \mathbf{r}_{\ell}\| \;\le\; \rho_{i\ell},
    \quad \ell \in \mathcal{N}_i^{\text{R}},
    \label{eq:robot-robot-upper}
\end{equation}
Although lower bounds for inter-robot measurements can also be obtained, our proposed formulation relies exclusively on the upper bounds to guarantee convexity. Geometrically, these bounds define convex inter-robot spherical regions. Because these pairwise geometric constraints are the primary source of coupling within the network, achieving decentralized cooperative localization requires decoupling them.  To accomplish this, the constraints are reformulated into Linear Matrix Inequalities (LMIs) and subsequently decomposed into resource-based local constraints.

\subsection{Objective}
Given the set of known landmark positions $\mathbf{B}_j$ and the bounded range measurements, the primary objective of this cooperative localization problem is to estimate the unknown 3D coordinates $\mathbf{r}_i \in \mathbb{R}^3$ and their corresponding shape matrix $\mathbf{P}_i$ for all robots $i \in \{1,\dots,N\}$. Conceptually, this is analogous to fitting a Maximum Volume Ellipsoid (MVE) within the bounded feasible region defined by the sensor measurements. In this framework, the center of the ellipsoid, denoted as $\mathbf{r}_i^c$, serves as the position estimate. Meanwhile, the optimized matrix $\mathbf{P}_i$ bounds the ellipsoid's volume, explicitly quantifying the spatial confidence of this estimate and serving as a direct mathematical analogue to the measurement covariance matrix provided by a standard GPS receiver. Consequently, this joint estimate ($\mathbf{r}_i^c$, $\mathbf{P}_i$) can be seamlessly utilized in downstream filtering pipelines to fuse with high-rate proprioceptive sensors (e.g., odometry or IMUs), acting as the absolute position correction mechanism in GPS-denied environments just as a GPS would.


Mathematically, let $\mathcal{E}_i$ denote the ellipsoid parameterized by its center $\mathbf{r}_i^c$ and shape matrix $\mathbf{P}_i$, formally defined as:
\begin{equation}
    \mathcal{E}_i = \left\{ \mathbf{r} \in \mathbb{R}^3 \;\middle|\; \mathbf{r} = \mathbf{P}_i \mathbf{u} + \mathbf{r}_i^c, \; \|\mathbf{u}\| \le 1 \right\},
    \label{eq:ellipsoid_def}
\end{equation}
where $\mathbf{u} \in \mathbb{R}^3$ is a bounded unit vector. Because the volume of the ellipsoid is proportional to $\det(\mathbf{P}_i)$, we minimized $-\log\det(\mathbf{P}_i)$ to find the position estimate. This entails solving the following optimization problem:
\begin{align*}
& \min_{\{\mathbf{r}_i^c, \mathbf{P}_i\}_{i=1}^N} \sum_{i=1}^N -\log\det(\mathbf{P}_i) \\
& \text{s.t.} \;\; \forall i \in \{1,\dots,N\}, \\
& \mathcal{E}_i \subseteq \left\{ \mathbf{r} \in \mathbb{R}^3 \;\middle|\; 
\begin{aligned}
& \underline{\rho}_{ij} \le \|\mathbf{r} - \mathbf{B}_j\| \le \bar{\rho}_{ij}, \; \forall j \in \mathcal{N}_i^{\text{A}} \\
& \|\mathbf{r}_i^c - \mathbf{r}^c_{\ell}\| \le \rho_{i\ell}, \; \forall \ell \in \mathcal{N}_i^{\text{R}}
\end{aligned}
\right\}.
\end{align*}

Furthermore, to address the spatial privacy requirements of sensitive operations, this optimizing problem must be solved in a fully decentralized manner. Rather than relying on a central server, each robot $i$ must estimate its own position $\mathbf{r}^c_i$ through local computation and restricted information exchange with its neighbors $\ell \in \mathcal{N}_i^{\text{R}}$, ensuring that its explicit spatial coordinates are never directly broadcasted.

\section{Construction of the Spatial Feasible Region}
In this section, we mathematically define the spatial region where each robot can potentially be located based on its sensor measurements. We build this feasible region in three progressive steps. First, we establish basic spherical regions using upper-bound range measurements from known landmarks. Second, we geometrically tighten these bounds by exploiting the lower-bound range measurements from those same landmarks. Finally, we enable cooperative localization by linking the fleet together using upper-bound inter-vehicle distance measurements. To guarantee that our localization algorithm can solve this problem efficiently, we formulate all of these spatial restrictions as convex Linear Matrix Inequalities (LMIs).

\subsection{Spherical Upper Bounds from Landmark Measurements}
\label{Upper-Bound-Range-Spheres}

\begin{figure}[htbp]
\centering
\includegraphics[width=0.85\linewidth]{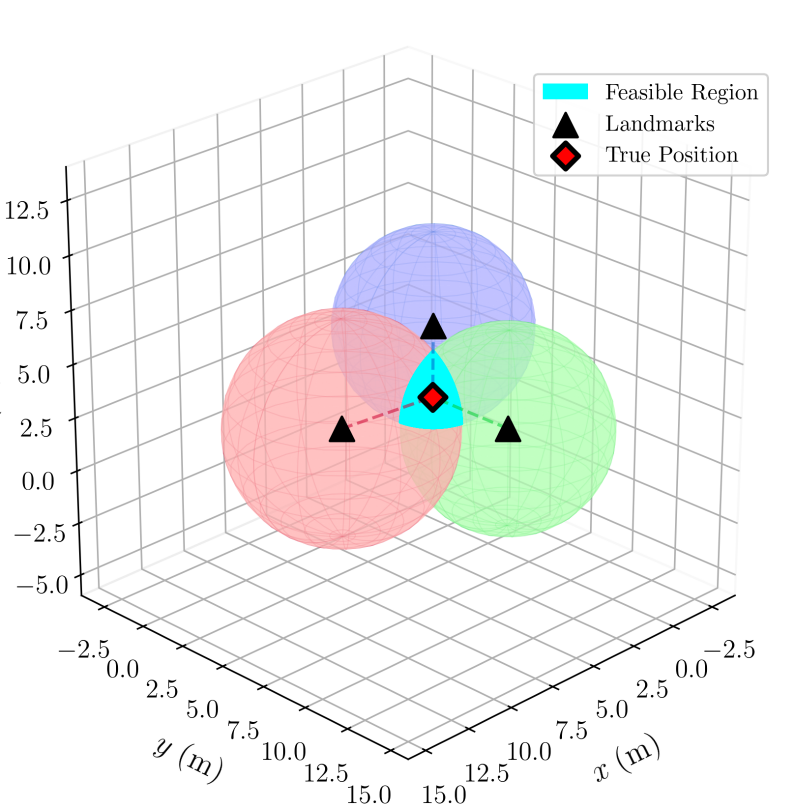}
\caption{Feasible region formed by the intersection of upper bound ($\bar\rho_{ij}$) range measurements from landmarks.}
\label{fig:outer-sphere-feasible-region}
\end{figure}

To construct the fundamental feasible region for each robot, we first evaluate the upper bounds of the range measurements obtained from the static landmarks. Geometrically, each upper-bound measurement $\bar{\rho}_{ij}$ defines a convex outer-sphere centered at the known landmark position $\mathbf{B}_j$. The true position of robot $i$ must logically reside within the intersection of these spherical regions (refer to Figure \ref{fig:outer-sphere-feasible-region}).  We formally define this feasible region as:
\begin{equation}
\mathcal{I}_i
=
\bigcap_{j \in \mathcal{N}_i^{\text{A}}}
\left\{
\mathbf{r} \in \mathbb{R}^3 \;\middle|\;
\|\mathbf{r} - \mathbf{B}_j\| \leq \bar{\rho}_{ij}
\right\},
\end{equation}

Our optimization objective is to inscribe the estimated uncertainty ellipsoid $\mathcal{E}_i$ entirely within this feasible space. Rather than grappling with the complex, composite geometry of $\mathcal{I}_i$ directly, we enforce containment on a per-landmark basis. Because enclosing an ellipsoid inside a sphere mathematically requires bounding one quadratic form by another, we apply the S-procedure and Schur complement \cite{rs17152637,boyd2004convex}. This allows us to reformulate the continuous spatial condition ($\mathcal{E}_i \subseteq \{\|\mathbf{r}_i - \mathbf{B}_j\| \leq \bar{\rho}_{ij}\}$) into a highly tractable Linear Matrix Inequality (LMI):
\begin{equation}
\begin{bmatrix}
\bar{\rho}_{ij}-\lambda_{ij} & (\mathbf{r}_i^c - \mathbf{B}_j)^\top & \mathbf{0} \\
\mathbf{r}^c_i - \mathbf{B}_j & \bar{\rho}_{ij} I_3 &\mathbf{P}_i \\
\mathbf{0} &\mathbf{P}_i & \lambda_{ij} I_3
\end{bmatrix} \succeq 0,\quad \exists \; \lambda_{ij} \ge 0.
\label{eq:basic-LMI}
\end{equation}
Enforcing \eqref{eq:basic-LMI} for a single landmark ensures the ellipsoid remains within that specific sphere. Therefore, simultaneously enforcing this LMI for all available landmarks ($\forall j \in \mathcal{N}^A_i$) guarantees that $\mathcal{E}_i$ is perfectly inscribed within the intersection $\mathcal{I}_i$.


\subsection{Feasible Region Tightening via Intersection Planes}
\label{sec:intersection-plane}


While the previous section established a baseline feasible region using upper-bound range measurements, we can significantly tighten this space by leveraging lower-bound measurements ($\underline{\rho}_{ij}$). These lower bounds naturally define exclusion zones—regions falling within the minimum separation distance of a landmark where the robot logically cannot reside. Since the exact feasible region defined by these bounds is inherently non-convex, we propose a convex relaxation utilizing intersection-plane constraints. 

\begin{figure}[htbp]
\centering
\includegraphics[width=0.85\linewidth]{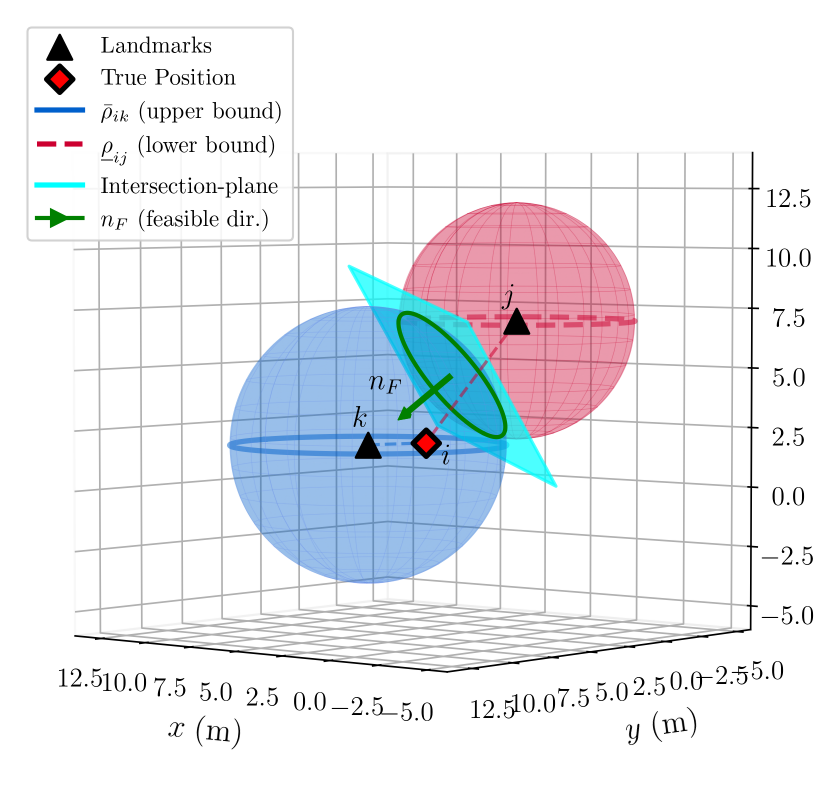}
\caption{Representation of the intersection plane constraint formed by measurements from two distinct landmarks.}
\label{fig:inner-sphere-feasible-region}
\end{figure}

\subsubsection{Intersection-plane Constraints}

Consider a scenario where robot $i$ processes measurements to two distinct landmarks, subjecting its spatial coordinate $\mathbf{r}_i$ to a lower-bound constraint with respect to landmark $\mathbf{B}_j$ ($\|\mathbf{r}_i - \mathbf{B}_j\| \geq \underline{\rho}_{ij}$) and an upper-bound constraint with respect to landmark $\mathbf{B}_k$ ($\|\mathbf{r}_i - \mathbf{B}_k\| \leq \bar{\rho}_{ik}$).


Geometrically, the feasible location of robot $i$ is constrained to lie outside the lower-bound exclusion sphere centered at $\mathbf{B}_j$, and inside the upper-bound bounding sphere centered at $\mathbf{B}_k$ (Figure \ref{fig:inner-sphere-feasible-region}). 

However, the boundary where these two spherical surfaces meet naturally defines an \emph{intersection plane}.  By calculating this plane and using it as a linear cut against the outer bounding sphere, we cleanly slice away the non-convex exclusion zone. 

\subsubsection{Derivation of the Intersection Plane}

Consider the squared-distance inequalities:
\begin{align}
\|\mathbf{r}_i - \mathbf{B}_j\|^2 &\geq \underline{\rho}_{ij}^2, \\
\|\mathbf{r}_i - \mathbf{B}_k\|^2 &\leq \bar{\rho}_{ik}^2.
\end{align}
Expanding both:
\begin{align}
\|\mathbf{r}_i\|^2 - 2 \mathbf{B}_j^\top \mathbf{r}_i + \|\mathbf{B}_j\|^2 &\geq \underline{\rho}_{ij}^2, \label{outer_sphere}
\end{align}
\begin{align}
\|\mathbf{r}_i\|^2 - 2 \mathbf{B}_k^\top \mathbf{r}_i + \|\mathbf{B}_k\|^2 &\leq \bar{\rho}_{ik}^2. \label{inner_sphere}
\end{align}
Subtracting \eqref{outer_sphere} from \eqref{inner_sphere} yields
\begin{equation}
2(\mathbf{B}_j - \mathbf{B}_k)^\top \mathbf{r}_i \leq \bar{\rho}_{ik}^2 - \underline{\rho}_{ij}^2 + \|\mathbf{B}_j\|^2 - \|\mathbf{B}_k\|^2.
\end{equation}
The above inequality represents the intersection-plane constraint. Defining the normal vector and offset as
\begin{align}
    \mathbf{n}_{jk} &\triangleq \mathbf{B}_j - \mathbf{B}_k, \\
    \gamma_{jk} &\triangleq \frac{1}{2}\left(\bar{\rho}_{ik}^2 - \underline{\rho}_{ij}^2 + \|\mathbf{B}_j\|^2 - \|\mathbf{B}_k\|^2 \right),
\end{align}
the intersection-plane constraint becomes
\begin{equation}
    \mathbf{n}_{jk}^\top \mathbf{r}_i \leq \gamma.
\end{equation}
Thus, the feasible region for robot $i$ is restricted to the half-space defined by the plane $\{\mathbf{r}\in\mathbb{R}^3 \mid \mathbf{n}_{jk}^\top \mathbf{r} = \gamma_{jk}\}$ and the inequality $\mathbf{n}_{jk}^\top \mathbf{r}_i \leq \gamma_{jk}$.

\subsubsection{Conditions for Valid Intersection}

The two spheres must intersect (or be tangent) in order for this plane to define a meaningful restriction. Let $d_{jk} = \|\mathbf{B}_k - \mathbf{B}_j\|$ be the distance between landmark centers. A non-empty intersection requires
\begin{equation}
|\underline{\rho}_{ij} - \bar{\rho}_{ik}| \leq d_{jk} \leq \underline{\rho}_{ij} + \bar{\rho}_{ik}.\label{eq:intersection-condtion}
\end{equation}

We utilize this requirement to formulate $\mathcal{P}_i$, defined as the set of all valid landmark pairs $(j,k)$ available to robot $i$. A pair is included in $\mathcal{P}_i$ if and only if the robot possesses a lower-bound measurement to landmark $\mathbf{B}_j$, an upper-bound measurement to landmark $\mathbf{B}_k$, and their resulting spheres satisfy the intersection condition in \eqref{eq:intersection-condtion}. Constructing this set is computationally vital, as it filters out redundant measurements and explicitly isolates the pairs that actively tighten the feasible region.



\subsubsection{Semidefinite Formulation of Plane Constraints}


We now embed the intersection-plane constraint into the ellipsoidal localization framework. Recall that the ellipsoid for robot $i$ is $\mathcal{E}_i$ with center $\mathbf{r}_i^c$ and shape matrix $\mathbf{P}_i \succeq 0$. The plane constraint $\mathbf{n}_{jk}^\top \mathbf{r}_i \leq \gamma_{jk}$ is enforced for all $\mathbf{r}_i \in \mathcal{E}_i$ by requiring that the entire ellipsoid lies in the half-space:
\begin{align}
&\mathbf{n}_{jk}^\top (\mathbf{P}_i \mathbf{u} + \mathbf{r}_i^c)  \leq \gamma_{jk} \quad \forall \;\mathbf{u}: \|\mathbf{u}\| \leq 1, \\
\iff&\mathbf{n}_{jk}^\top\mathbf{P}_i \mathbf{u}  \leq \gamma_{jk} - \mathbf{n}_{jk}^\top\mathbf{r}_i^c \quad \forall \;\mathbf{u}: \|\mathbf{u}\| \leq 1, \\
\iff&\|\mathbf{P}_i^\top\mathbf{n}_{jk}\|  \leq \gamma_{jk} - \mathbf{n}_{jk}^\top\mathbf{r}_i^c.
\end{align}
By applying the Schur complement, this constraint can be equivalently expressed as the following Linear Matrix Inequality (LMI):
\begin{equation}
\begin{bmatrix}
\gamma_{jk} - \mathbf{n}_{jk}^\top \mathbf{r}_i^c & \mathbf{n}_{jk}^\top\mathbf{P}_i \\
\mathbf{P}_i^\top \mathbf{n}_{jk} & (\gamma_{jk} - \mathbf{n}_{jk}^\top \mathbf{r}_i^c) I_3
\end{bmatrix} \succeq 0. \label{inner-sphere-lmis}
\end{equation}


\subsection{Enabling Cooperation via Inter-Robot Coupling}

Building upon the tightened, landmark-based feasible regions, we now integrate inter-robot range measurements ($\rho_{i\ell}$) as spatial coupling constraints to enable true cooperative localization.


Suppose robots $i$ and $\ell$ are connected by an outer bound on their distance, denoted by ${\rho}_{i\ell}$. We require
\begin{equation}
{\rho}_{i\ell} \ge \|\mathbf{r}^c_i - \mathbf{r}^c_\ell\|, 
\label{eq:coupling-const}
\end{equation}
where $\mathbf{r}^c_i$ and $\mathbf{r}^c_\ell$ denote the (center) position variables of the ellipsoids for robots $i$ and $\ell$, respectively. Squaring both sides gives
\begin{equation}
{\rho}_{i\ell}^2 - (\mathbf{r}^c_i - \mathbf{r}^c_\ell)^\top (\mathbf{r}^c_i - \mathbf{r}^c_\ell) \ge 0.\label{eq:coupling-const1}
\end{equation}
By the Schur complement, this is equivalent to the LMI
\begin{equation}
\begin{bmatrix}
{\rho}_{i\ell} & (\mathbf{r}^c_i - \mathbf{r}^c_\ell)^\top \\
\mathbf{r}^c_i - \mathbf{r}^c_\ell & {\rho}_{i\ell} I_3
\end{bmatrix} \succeq 0.
\label{eq:intervehicle-lmi}
\end{equation}
This transformation provides a standard, strictly convex representation of the inter-robot coupling constraints. Note that the ordering of the indices $i$ and $\ell$ is arbitrary; swapping them merely negates the off-diagonal block in \eqref{eq:intervehicle-lmi}, which preserves the matrix's positive semidefiniteness and identically recovers the squared distance bound in \eqref{eq:coupling-const1}.

\section{The Privacy-Preserving Decentralized Cooperative Localization Method}

This section presents the core mathematical architecture of our privacy-preserving localization method. To provide a rigorous benchmark for our approach, we begin by defining a Centralized Oracle (CO) that jointly estimates the Maximum-Volume Inscribed Ellipsoids (MVE) for the entire fleet using globally aggregated data. Recognizing the severe scalability and privacy limitations of such an architecture, we then formulate our Decentralized Cooperative Localization (DCL) framework. We detail how the centralized optimization problem is systematically decomposed into localized Semi-Definite Programs (SDPs), allowing the fleet to cooperatively refine their spatial bounds through a strictly anonymized dual-variable consensus protocol.

\subsection{Centralized Oracle (CO)}
Representing the theoretical optimal baseline for our network, this approach utilizes an SDP to jointly estimate the positions and uncertainty ellipsoids for all robots simultaneously. By aggregating all available data at a central node, this formulation comprehensively incorporates the upper-bound spherical constraints \eqref{eq:basic-LMI}, the intersection-plane constraints \eqref{inner-sphere-lmis}, and the inter-robot coupling constraints \eqref{eq:intervehicle-lmi}. 

The optimization problem is:
\begin{align}
\min_{\{\mathbf{r}_i^c,\mathbf{P}_i\succeq 0\}} \quad & \sum_{i=1}^N - \log \det(\mathbf{P}_i) \\[0.2em]
\text{s.t.} \quad & \eqref{eq:basic-LMI}, \quad \forall i, \; \forall j \in \mathcal{N}^A_i, \label{eq:central-sphere-lmi} \\ 
&\eqref{inner-sphere-lmis}, \quad \forall i, \; \forall (j,k) \in \mathcal{P}_i , \label{eq:central-inter-lmi}\\
&\eqref{eq:intervehicle-lmi}, \quad \forall (i,\ell)\in\mathcal{E}. \label{eq:central-coupling}
\end{align}
Here, $\mathcal{E}$ denotes the set of all robot pairs $(i,\ell)$ that are within mutual sensing range. Because the optimization solver has simultaneous access to the entire network's state variables, this centralized SDP provides the optimal performance benchmark for our comparative analysis. However, deploying this centralized architecture in real-world scenarios relies on a single computational hub to continuously aggregate raw telemetry and broadcast optimized solutions, raising severe scalability and privacy concerns.

\subsection{Decentralized Cooperative Localization (DCL)}
This approach represents our proposed privacy-preserving formulation. The iterative decentralized architecture mirrors the mathematical rigor of the centralized formulation but intentionally splits the inter-robot dependencies to enable local computation and restricted neighbor-to-neighbor communication. 

To handle inter-robot spatial relationships without violating privacy, the global coupling constraint \eqref{eq:intervehicle-lmi} must be explicitly decomposed so that explicit spatial coordinates ($\mathbf{r}_i^c$) are never broadcasted. By introducing an auxiliary matrix variable $\mathcal{R}_{i\ell} = \mathcal{R}_{\ell i} \in \mathbb{R}^{4 \times 4}$ and  $\mathcal{M}_{i\ell}(\mathbf{r}_i^c) = \begin{bmatrix}
{\rho}_{i\ell} & 2{\mathbf{r}_i^c}^\top \\
2\mathbf{r}^c_i & {\rho}_{i\ell} I_3
\end{bmatrix}$, we reformulate the coupling into two independent local constraints:
\begin{align}
&\begin{cases}
\mathcal{M}_{i\ell}(\mathbf{r}_i^c) + \mathcal{R}_{i\ell} \succeq 0, & \text{if } i<\ell,\\
\mathcal{M}_{i\ell}(-\mathbf{r}_i^c) - \mathcal{R}_{i\ell} \succeq 0, & \text{if } i>\ell,
\end{cases} \label{eq:split-j}\\[0.5em]
&\begin{cases}
    \mathcal{M}_{i\ell}(-\mathbf{r}_\ell^c) -\mathcal{R}_{\ell i} \succeq 0 & \text{if } i<\ell, \\
    \mathcal{M}_{i\ell}(\mathbf{r}_\ell^c) +\mathcal{R}_{\ell i} \succeq 0 & \text{if } i>\ell.
\end{cases} \label{eq:split-k}
\end{align}

Algebraically summing \eqref{eq:split-j} and \eqref{eq:split-k} recovers the original coupling constraint. Consequently, this formulation serves as a sufficient decomposition of \eqref{eq:intervehicle-lmi}. The indexing conditions ($i < \ell$ versus $i > \ell$) serve as a strict symmetry-breaking mechanism for the undirected communication edges. By enforcing a lexicographical order, we ensure that one robot along the edge incorporates the shared resource matrix $\mathcal{R}_{i\ell}$ with a positive sign, while its neighbor incorporates it with a negative sign.


At each iteration, robot $i$ solves the following local SDP:
\begin{align}
\min_{\mathbf{r}_i^c,\mathbf{P}_i} \quad &
   - \log \det(\mathbf{P}_i)  \label{eq:dec-obj} \\
\text{s.t.} \quad &
\eqref{eq:basic-LMI}, \quad \forall j \in \mathcal{N}^A_i, \label{eq:dec-local1}\\ 
&\eqref{inner-sphere-lmis}, \quad \forall (j,k) \in \mathcal{P}_i,  \label{eq:dec-local2}\\ 
&\eqref{eq:split-j}, \quad \forall \ell \in \mathcal{N}^R_i \label{eq:dec-coupling} 
\end{align}

After solving the local SDP, robot $i$:
\begin{enumerate}
    \item Exchanges its dual variable matrix $\Lambda_{i\ell}$ corresponding to equations \eqref{eq:dec-coupling} with its neighbors $\ell \in \mathcal{N}^R_i$.
    \item Updates the shared matrices $\mathcal{R}_{i\ell}$ using a step size $\alpha$: 
    \begin{align}
    \begin{cases}
    \mathcal{R}_{i\ell} \leftarrow \mathcal{R}_{i\ell} - \alpha (\Lambda_{i\ell} - \Lambda_{\ell i}), & \text{if } i<\ell \\
    \mathcal{R}_{i\ell} \leftarrow \mathcal{R}_{i\ell} + \alpha (\Lambda_{i\ell} - \Lambda_{\ell i}), & \text{if } i>\ell
    \end{cases},~~\forall \ell \in \mathcal{N}^R_i.\label{eq:update-resource}
    \end{align}
\end{enumerate}

By initializing the system such that $\mathcal{R}_{i\ell} = \mathcal{R}_{\ell i}$, the complementary signs in the update rule \eqref{eq:update-resource} guarantee that the shared matrices remain perfectly consistent across the two robots for all future iterations.

Theoretically, while summing the split constraints \eqref{eq:split-j} and \eqref{eq:split-k} algebraically recovers the original coupled constraint \eqref{eq:intervehicle-lmi}, demanding the existence of this specific auxiliary matrix $\mathcal{R}_{i\ell}$ mathematically restricts the optimization space. Consequently, this decoupled formulation acts as a sufficient condition, firmly establishing the Centralized Oracle as a strict theoretical lower bound.

\textit{Crucially, this specific decomposition enables a strictly privacy-preserving framework. Instead of broadcasting sensitive positional data, robots solve localized SDPs and achieve consensus on $\mathcal{R}_{i\ell}$ solely by sharing abstract dual variables. With each successive iterative update step, the DCL algorithm systematically closes the optimality gap introduced by this sufficient condition, driving the network toward the true centralized benchmark while maintaining strict spatial privacy.}

\subsection{Theoretical Justification via Subgradient Primal Decomposition}

This subsection mathematically validates the DCL framework. We first define the localized value functions to construct a global master problem. We then correlate this master problem to the Centralized Oracle (CO) and demonstrate how subgradient descent systematically utilizes local computations to drive the decentralized network toward the centralized theoretical lower bound.

\subsubsection{Local Value Functions and the Global Master Problem}
To decouple the network, we fix the collection of edge matrices incident to robot $i$ as $\mathcal{R}_i := \{\mathcal{R}_{i\ell}\}_{\ell\in\mathcal{N}_i^R}$. We define the optimal local cost (value function) for each robot as:
\begin{equation}
\phi_i(\mathcal{R}_i) := \min_{\mathbf{r}_i^c, \mathbf{P}_i \succeq 0} \;\; -\log\det(\mathbf{P}_i)
\label{eq:value-function}
\end{equation}
subject to all local constraints \eqref{eq:dec-local1}, \eqref{eq:dec-local2}, and the split coupling limits \eqref{eq:dec-coupling}. 

From these independent local subproblems, the network-level optimization can be cleanly reconstructed as the global \emph{master problem}:
\begin{equation}
\min_{\{\mathcal{R}_{i\ell}\}_{(i,\ell)\in\mathcal{E}}} \;\; \Phi(\mathcal{R}) := \sum_{i=1}^N \phi_i(\mathcal{R}_i).
\label{eq:master-problem}
\end{equation}

\subsubsection{Correlating the Master Problem to the Centralized Oracle (CO)}
The global master problem and the Centralized Oracle (CO) share the exact same global objective function. Furthermore, the master problem strictly preserves all absolute landmark measurement constraints (\eqref{eq:central-sphere-lmi} and \eqref{eq:central-inter-lmi}) exactly as they appear in the CO. 

The formulations differ only in their treatment of inter-robot spatial relationships. Rather than enforcing the original relative coupling \eqref{eq:central-coupling} found in the CO, the global master problem relies on the split constraints \eqref{eq:split-j} and \eqref{eq:split-k} for all edges $(i,\ell) \in \mathcal{E}$. Enforced via a lexicographical ordering ($i < \ell$), these split constraints act as a strict sufficient condition for the original coupling \eqref{eq:central-coupling}.

Because the only structural difference is the introduction of the auxiliary matrix $\mathcal{R}_{i\ell}$, the global master problem inherently imposes an extra restriction on the optimization space. Consequently, while the objectives and local constraints are identical, this restricted feasible space makes the global master problem a conservative approximation, establishing the CO as its strict theoretical lower bound.

\subsubsection{Closing the Optimality Gap via Subgradient Descent}
To systematically drive the decentralized network toward the theoretical performance limit of the CO, the algorithm must iteratively minimize the master objective $\Phi(\mathcal{R})$. We achieve this via subgradient descent, where solving the localized subproblems directly yields the exact mathematical components needed to formulate the subgradient for the global master problem.

Consider an edge $(i,\ell)\in\mathcal{E}$ with $i< \ell$. Let $\Lambda_{i\ell}$ be the optimal dual multiplier of robot $i$'s edge constraint in its local SDP ($\phi_i(\mathcal{R}_i)$), and let $\Lambda_{\ell i}$ be the corresponding optimal dual multiplier in robot $\ell$'s local SDP ($\phi_\ell(\mathcal{R}_\ell)$).

Under Slater's condition for each local SDP---which naturally holds because our bounded measurement noise ensures the robot's true physical position resides in the strictly feasible interior---strong duality applies. For any perturbation $\Delta$ of $\mathcal{R}_{i\ell}$, evaluating the optimal local costs using their Lagrangian duals yields:
\begin{align}
\phi_i(\mathcal{R}_{i\ell}+\Delta) &\ge \phi_i(\mathcal{R}_{i\ell}) + \langle \Lambda_{i\ell}, \Delta\rangle, \nonumber \\
\phi_\ell(\mathcal{R}_{i\ell}+\Delta) &\ge \phi_\ell(\mathcal{R}_{i\ell}) - \langle \Lambda_{\ell i}, \Delta\rangle,
\label{eq:subgrad-ineq}
\end{align}
where the minus sign in the second inequality occurs because $\mathcal{R}_{i\ell}$ enters robot $\ell$'s constraint with the opposite sign (cf. \eqref{eq:split-k}).

Thus, from \eqref{eq:subgrad-ineq}, the local duals act as the subgradients of the local value functions: $\Lambda_{i\ell} \in \partial_{\mathcal{R}_{i\ell}}\phi_i$ and $-\Lambda_{\ell i} \in \partial_{\mathcal{R}_{i\ell}}\phi_\ell$. Because the global master objective is additively separable ($\Phi = \sum_i \phi_i$), summing these locally derived components gives the exact subgradient for the global master problem:
\begin{equation}
\Lambda_{i\ell} - \Lambda_{\ell i} \in \partial_{\mathcal{R}_{i\ell}}\Phi(\mathcal{R}), \qquad \forall (i,\ell)\in\mathcal{E} , i<\ell.
\label{eq:master-subgradient}
\end{equation}

\subsubsection{ The DCL Primal Update}
Executing a subgradient descent step on the master problem \eqref{eq:master-problem} using the derived subgradient \eqref{eq:master-subgradient} yields:
\begin{equation}
\mathcal{R}_{i\ell}^{(t+1)} = \mathcal{R}_{i\ell}^{(t)} - \alpha_t\left(\Lambda_{i\ell}^{(t)} - \Lambda_{\ell i}^{(t)}\right), \qquad \forall (i,\ell)\in\mathcal{E}, i<\ell,
\end{equation}
which is exactly the DCL primal update stated previously. The identical derivation holds symmetrically for the case when $i>\ell$. Through these iterative dual-variable exchanges, the network systematically traverses the subgradient, continuously shrinking the optimality gap introduced by the sufficient condition.

\section{Simulation Results}

All optimization-based methods were implemented in Julia using JuMP for model formulation and MOSEK as the SDP solver. We evaluate our approaches through extensive 3D Monte Carlo simulations. The primary goal of these evaluations is to demonstrate that our proposed DCL framework generates highly accurate, divergence-free absolute position estimates. 

\subsection{Baselines}
To rigorously evaluate our proposed DCL framework, we establish the following two baseline methods.
\subsubsection{Spherical Bounds (SB)}
This baseline approach, adopted from \cite{rs17152637}, relies exclusively on the upper-bound spherical constraints \eqref{eq:basic-LMI} derived from landmark measurements. For robot $i$, we solve:
\begin{align*}
\min_{\mathbf{r}_i^c,\,\mathbf{P}_i \succeq 0} \quad & -\log \det\mathbf{P}_i \\
\text{s.t.} \quad &
(\ref{eq:basic-LMI}) , \quad \;\forall j \in \mathcal{N}^A_i.
\end{align*}

\subsubsection{Spherical Bounds and Planar Bounds (SB+PB)}
This intermediate formulation incorporates both the upper-bound spherical constraints \eqref{eq:basic-LMI} and the intersection-plane constraints \eqref{inner-sphere-lmis} derived from static landmarks. For a given robot $i$, the localized SDP is formulated as:
\begin{align*}
\min_{\mathbf{r}_i^c,\,\mathbf{P}_i \succeq 0} \quad & -\log \det\mathbf{P}_i \\
\text{s.t.} \quad &
\eqref{eq:basic-LMI} , \; \;\forall j \in \mathcal{N}^A_i,
\text{  and  } \eqref{inner-sphere-lmis} \; \;\forall (j,k) \in \mathcal{P}_i.
\end{align*}

\subsection{Simulation Setup}

We simulate a network of $N = 10$ mobile robots and $M$ static landmarks ($15 \leq M \leq 20$). Each robot is equipped with a sensor capable of a maximum range of $R_s = 50\,\mathrm{m}$. To mirror realistic sensing conditions, we construct the actual range measurements acquired by the robots by corrupting the ground-truth Euclidean distances with maximum additive noise margins. Specifically, the robots process the upper-bound measurement $\bar{\rho}_{ij}$ and lower-bound measurement $\underline{\rho}_{ij}$ to each landmark, as well as the upper bound inter-robot distance measurement $\rho_{i\ell}$ to each neighbor, generated as follows:
\begin{align*}
\bar{\rho}_{ij} = d_{ij} + \sigma_o, \quad \underline{\rho}_{ij} =  d_{ij} - \sigma_i, \quad \rho_{i\ell} = d_{i\ell} + \sigma_c.
\end{align*}
Here, $d_{ij}$ and $d_{i\ell}$ are the ground-truth distances to landmark $j$ and robot $\ell$, and $\sigma_o=$ $\sigma_i=$  $\sigma_c = 0.2$  represent the predefined  additive noise margins, respectively.

We evaluate performance across 100 random trials, measuring the absolute position error $e_i = \| {\mathbf{r}}^c_i - \mathbf{r}_i^* \|_2$ and the total computation time. Between trials, the positions of both the landmarks and robots are randomized, subject to a connectivity constraint ensuring that each robot connects to at least three neighbors ($|\mathcal{N}^R_i| \geq 3, \; \forall i$).

\subsection{Results}
\begin{figure}[htbp]
\centering
\includegraphics[width=0.85\linewidth]{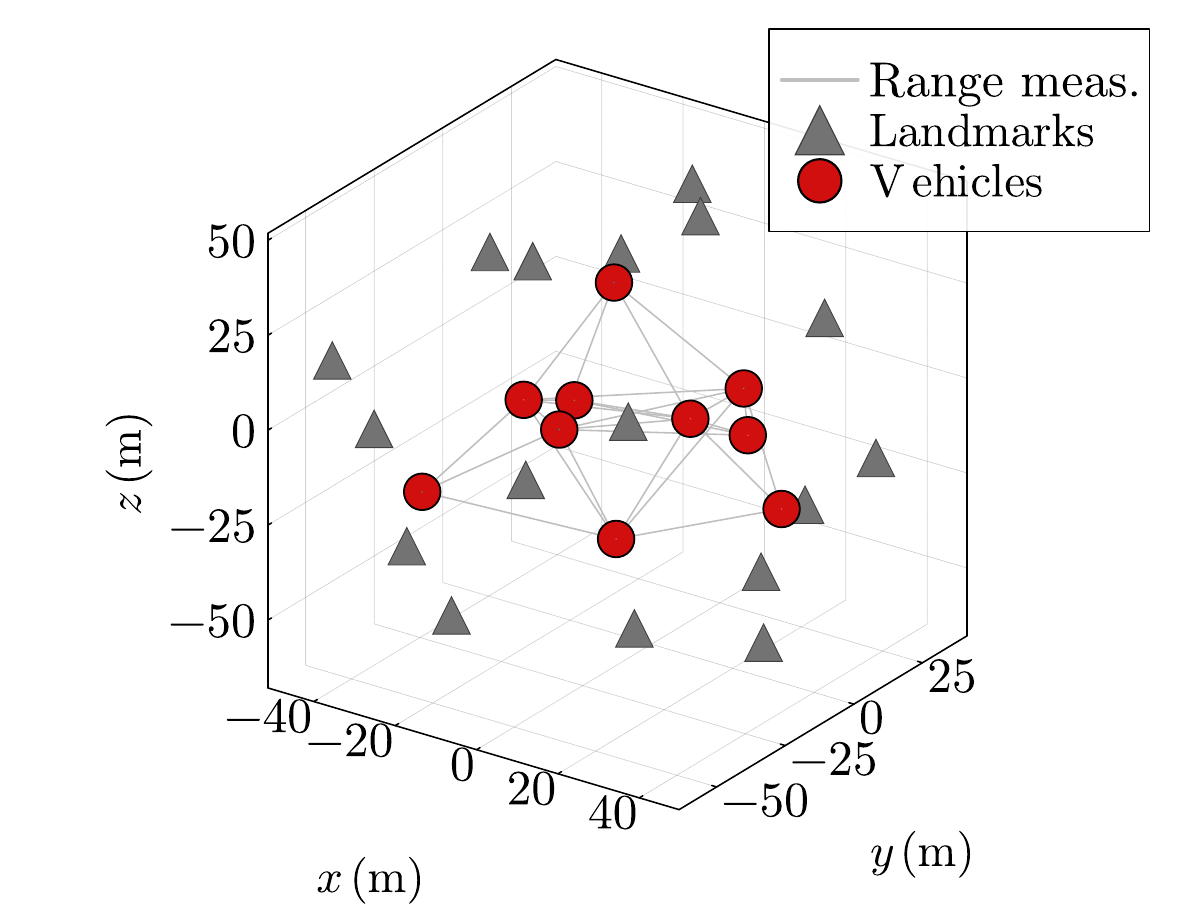}
\caption{Representative network topology.}
\label{fig:network}
\end{figure}

\begin{figure}[t]
    \centering
    \begin{subfigure}[b]{0.48\linewidth}
        \centering
        \includegraphics[width=\linewidth]{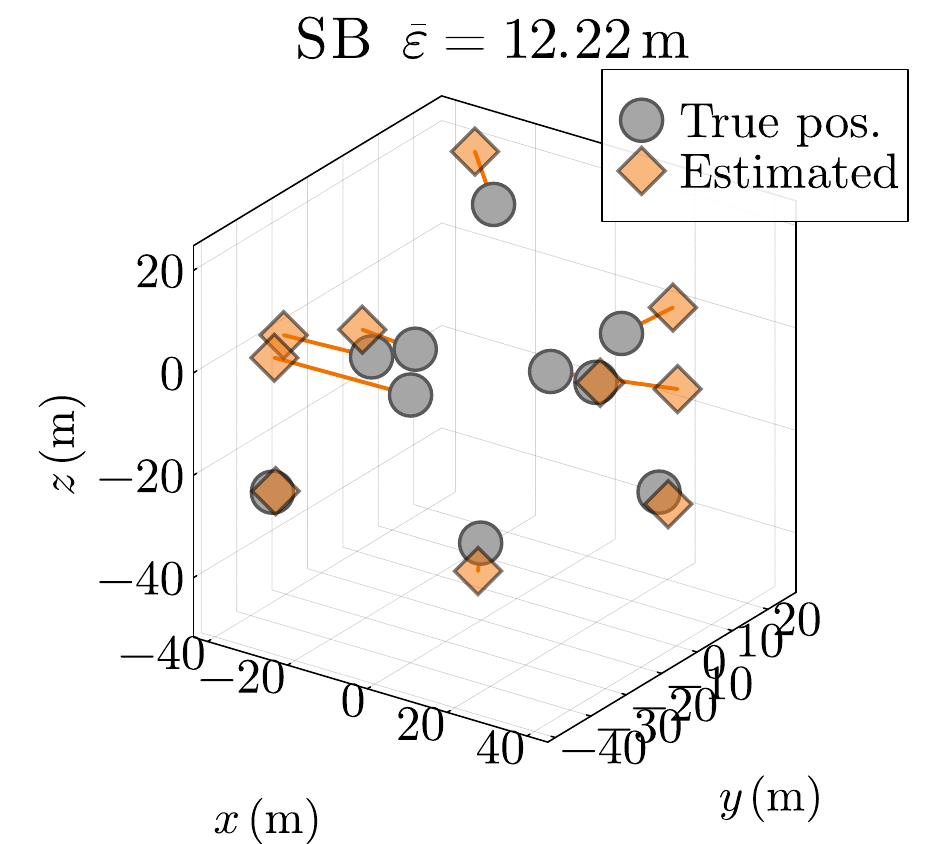}
        \caption{ Spherical Bounds (Baseline method) }
        \label{fig:sub1}
    \end{subfigure}
    \hfill
    \begin{subfigure}[b]{0.48\linewidth}
        \centering
        \includegraphics[width=\linewidth]{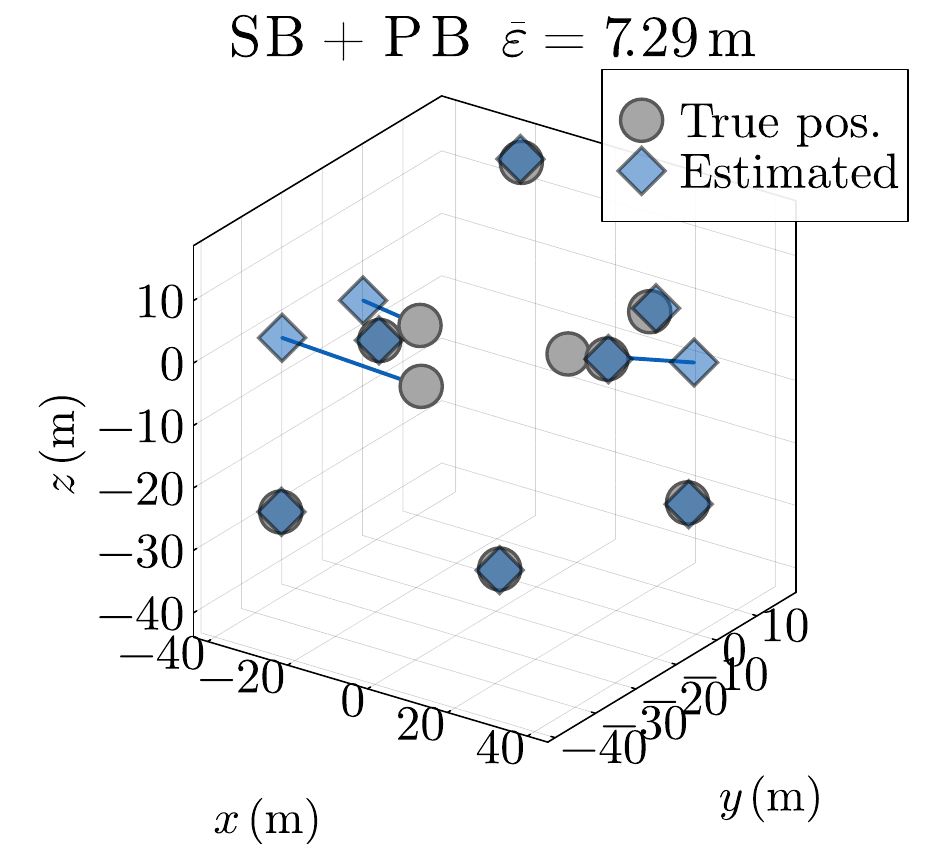}
        \caption{Spherical Bounds + Planar Bounds}
        \label{fig:sub2}
    \end{subfigure}

    \vspace{1ex}

    \begin{subfigure}[b]{0.48\linewidth}
        \centering
        \includegraphics[width=\linewidth]{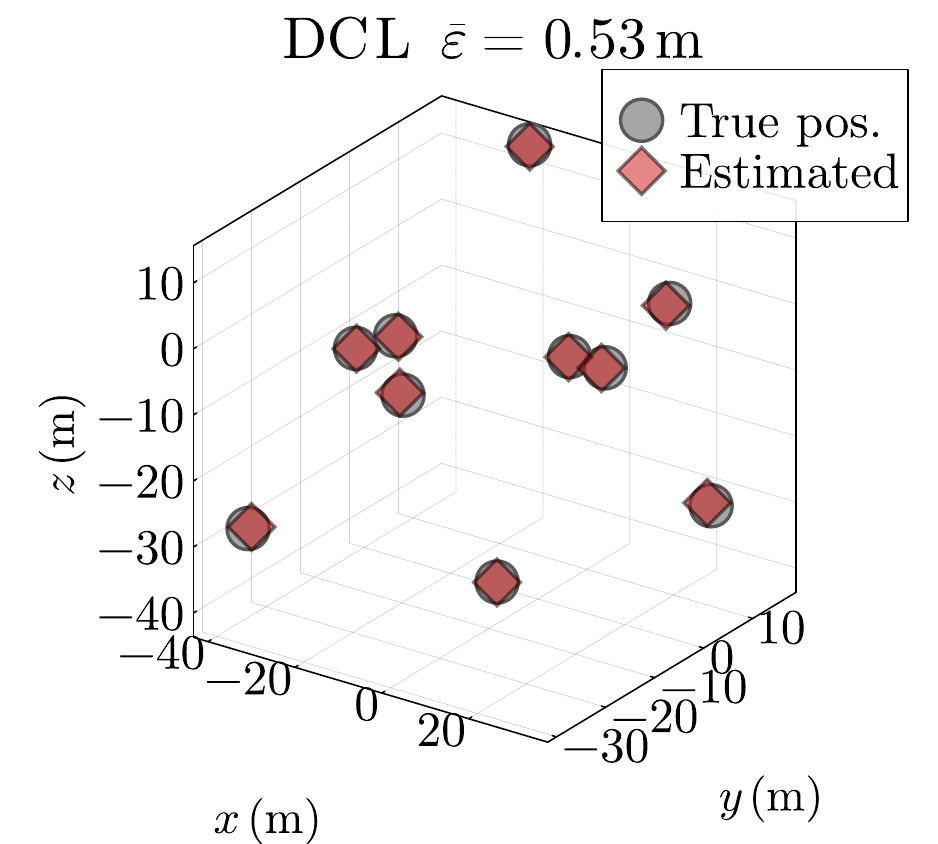}
        \caption{Decentralized Cooperative Localization}
        \label{fig:sub3}
    \end{subfigure}
    \hfill
    \begin{subfigure}[b]{0.48\linewidth}
        \centering
        \includegraphics[width=\linewidth]{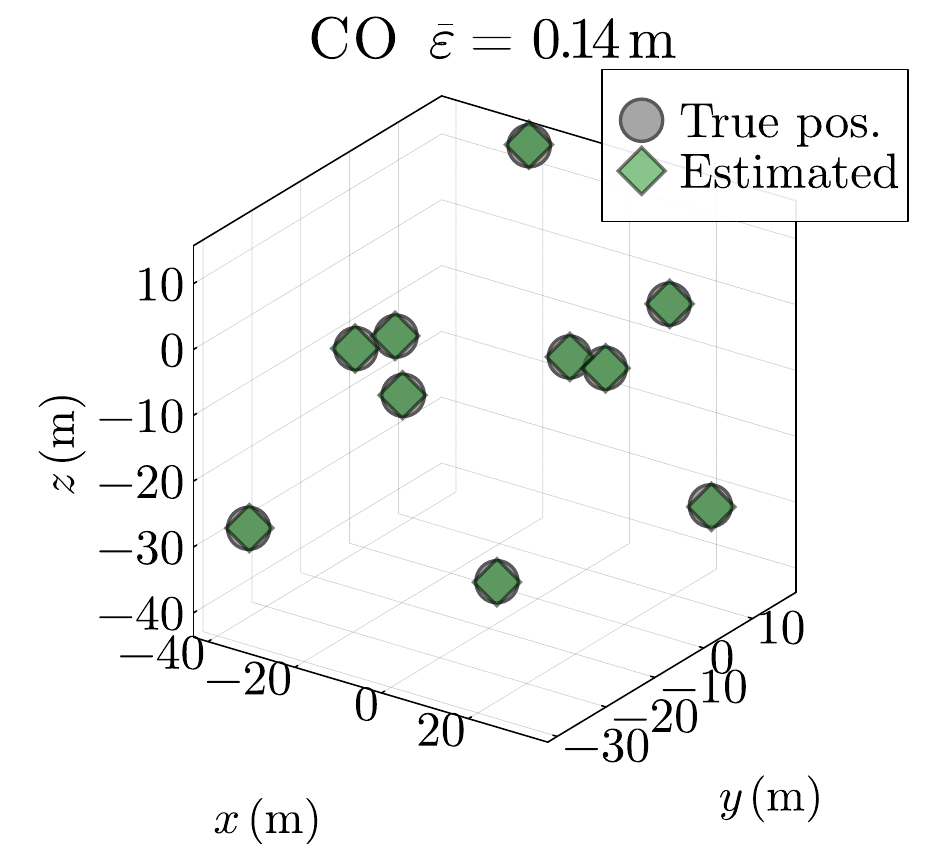}
        \caption{Centralized Oracle}
        \label{fig:sub4}
    \end{subfigure}

    \caption{Estimates vs True Position}
    \label{fig:positions}
\end{figure}

Fig.~\ref{fig:network} illustrates a representative simulation topology featuring randomly distributed landmarks and dense inter-robot connectivity. The corresponding position estimates, shown in Fig.~\ref{fig:positions}, highlight the progressive accuracy improvements achieved by our proposed formulations. While the standard spherical baseline (Fig.~\ref{fig:sub1}) suffers from pronounced displacement errors, incorporating local intersection-plane bounds significantly reduces these inaccuracies (Fig.~\ref{fig:sub2}). However, in this uncoupled formulation, robots with insufficient local landmark measurements still exhibit severe localization uncertainty. By leveraging inter-robot coupling constraints, our proposed Decentralized Cooperative Localization method (Fig.~\ref{fig:sub3}) effectively compensates for this sparse local data. It tightly bounds the true positions, successfully correcting the estimates of previously uncertain robots, and virtually closes the performance gap with the centralized oracle (Fig.~\ref{fig:sub4}).

\begin{figure}[t]
    \centering
    \begin{subfigure}[b]{0.48\linewidth}
        \centering
        \includegraphics[width=\linewidth]{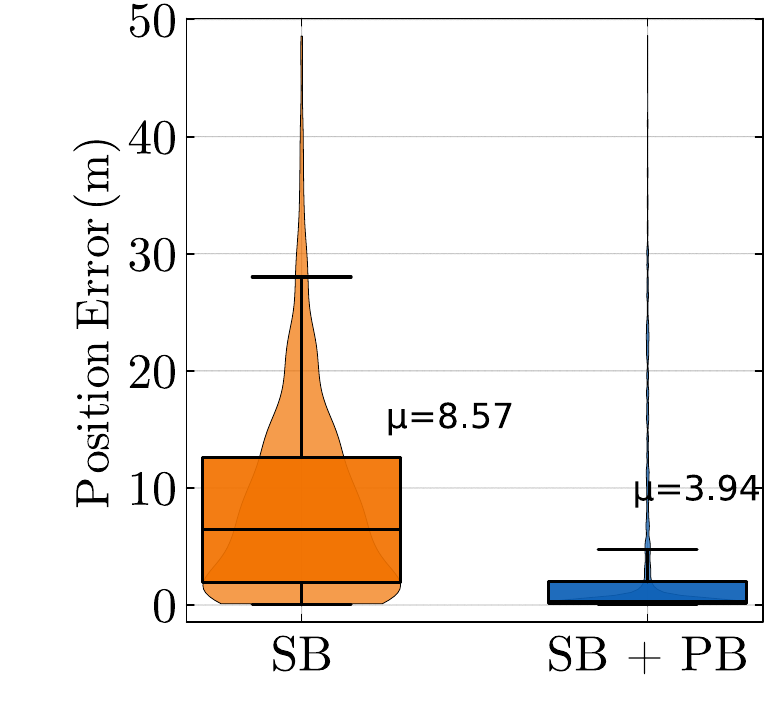}
        \caption{Comparison between SB and SB + PB}
        \label{fig:fig_violin_outer_dec}
    \end{subfigure}
    \hfill
    \begin{subfigure}[b]{0.48\linewidth}
        \centering
        \includegraphics[width=\linewidth]{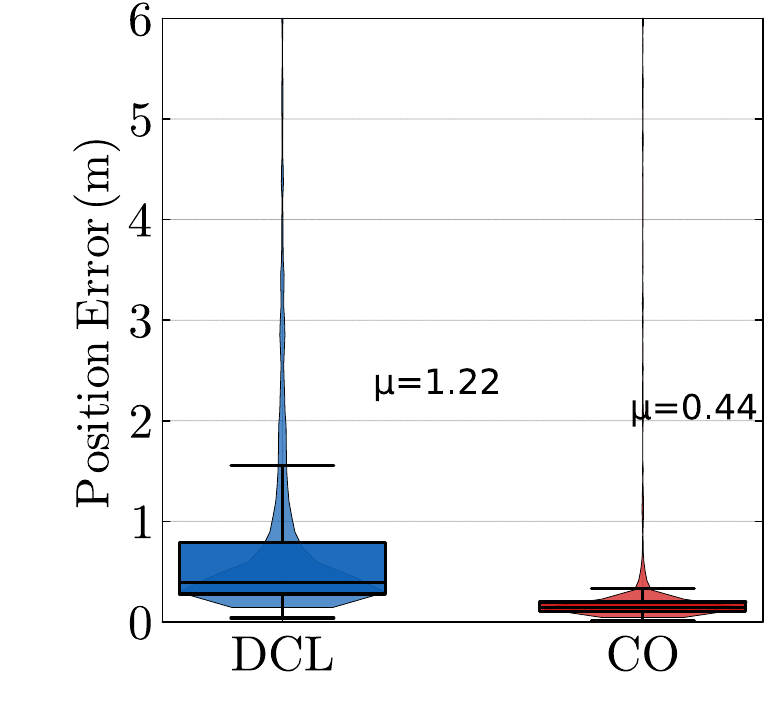}
        \caption{Comparison between DCL and CO}
        \label{fig:fig_violin_dec_prop_cent}
    \end{subfigure}
    \caption{Error distribution for 4 methods (100 trials).}
    \label{fig:violin}
    
\end{figure}

\begin{figure}[t]
    \centering
    \begin{subfigure}[b]{0.48\linewidth}
        \centering
        \includegraphics[width=\linewidth]{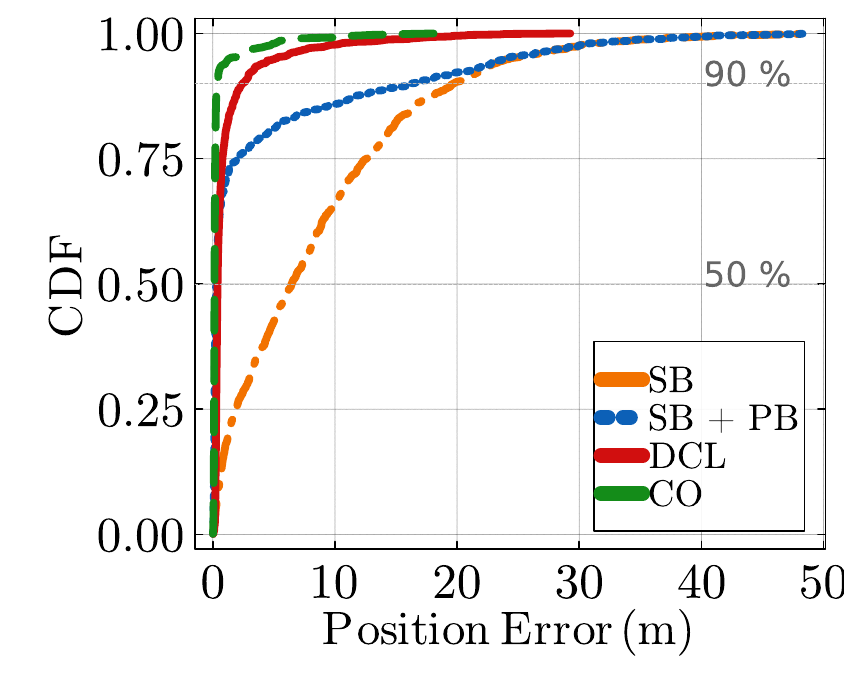}
        \caption{Error CDF reveals that DCL achieves CO-quality solutions while maintaining  privacy. }
        \label{fig:cdf}
    \end{subfigure}
    \hfill
    \begin{subfigure}[b]{0.48\linewidth}
        \centering
        \includegraphics[width=\linewidth]{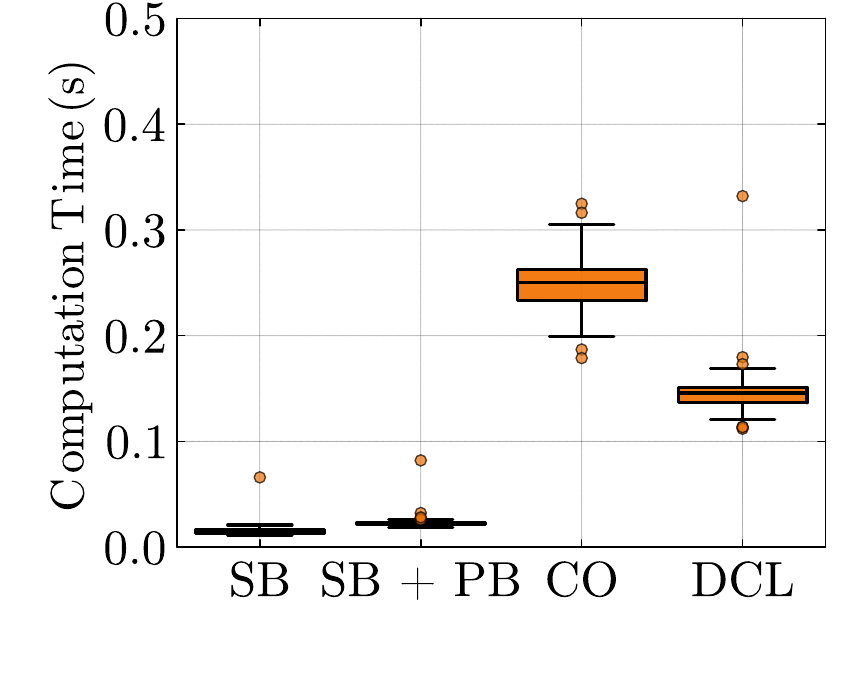}
        \caption{Computation time analysis shows that the proposed DCL method remains practical.}
        \label{fig:fig_time_box}
    \end{subfigure}
    \caption{Error and computational efficiency comparison.}
\end{figure}

This qualitative performance improvement is further substantiated quantitatively by the error distribution across the 100 randomized trials (Fig.~\ref{fig:violin}). While the isolated addition of intersection planes provides a clear, measurable benefit over the standard spherical baseline, the proposed Decentralized Cooperative Localization (DCL) method systematically minimizes the residual localization error. By leveraging inter-robot spatial coupling, DCL enhances network-wide accuracy, enabling robots with poor landmark visibility to iteratively correct their position estimates using relative distance data from well-localized peers. Consequently, the empirical Cumulative Distribution Function (CDF) of the positioning error (Fig.~\ref{fig:cdf}) demonstrates the robust effectiveness of the proposed methodology, confirming a significantly higher probability of bounding the error within stringent operational tolerances.

The DCL results presented herein were obtained using $K = 5$ consensus iterations with a parameter setting of $\alpha = 15$, yielding a highly efficient average computation time of $0.145\,\mathrm{s}$ (Figure \ref{fig:fig_time_box}). In our decentralized implementation, we initialize the shared resource matrices ($\mathcal{R}_{i\ell}$) with zero matrices before the first communication round. Because the network lacks initial consensus, starting from this zero-state can make the local Semi-Definite Programs (SDPs) overly restrictive and mathematically infeasible right at the start. To guarantee the solver always finds a valid solution, we introduce non-negative local slack variables directly into the inter-robot range measurements. By essentially adding a temporary flexible buffer to the measured distances, these variables relax the rigid spatial constraints just enough to safely bootstrap the optimization process.
While increasing the number of iterations improves localization accuracy, it inherently incurs a higher computational latency. Similarly, tuning the step size $\alpha$ dictates convergence dynamics: lower values guarantee stability at the cost of speed, whereas higher values accelerate updates but risk overshooting. Together, these parameters establish a clear, tunable trade-off between precision and real-time applicability, enabling the network to dynamically adapt to specific operational requirements.

\section{Conclusion}
In this paper, we presented a robust, privacy-preserving framework for Decentralized Cooperative Localization (DCL) using range-only measurements. By discarding  probabilistic noise models in favor of strictly bounded noise, we framed the localization problem as a convex Semi-Definite Programming (SDP) problem. 

The simulation results validate several key advantages of our framework. First, embedding geometric intersection-plane constraints immediately tightens feasible sets, improving even the baseline decentralized SDP. Second, exchanging dual variables effectively propagates positional corrections across the fleet without ever communicating sensitive coordinate data, inherently preserving operational privacy. Finally, because the local SDP dimensions rely solely on neighbor count rather than total fleet size, the proposed architecture scales favorably to massive robotic deployments.

Future work will focus on transitioning this framework from simulation to physical multi-robot hardware. Specifically, the integration of DCL outputs into standard Extended Kalman Filter (EKF) pipelines to fuse with high-rate proprioceptive sensors is planned. By providing robust, globally guaranteed correction anchors, the proposed privacy-preserving framework can serve as a direct, highly reliable substitute for GPS in unstructured and fully GPS-denied environments. Additionally, this EKF integration will resolve the vulnerabilities associated with high-speed motion and asynchronous communication. By relying on proprioceptive sensors for continuous high-rate tracking, the DCL consensus can function as a delayed absolute position update, ensuring stable localization without risking solver divergence from stale dual variables.  

\bibliographystyle{IEEEtran}
\bibliography{references}

@article{yu2023balancing,
  title={Balancing localization accuracy and location privacy in mobile cooperative localization},
  author={Yu, Dan and Shi, Xiufang and Chai, Li and Zhang, Wen-An and Chen, Jiming},
  journal={IEEE Transactions on Signal Processing},
  volume={71},
  pages={2804--2818},
  year={2023},
  publisher={IEEE}
}

@inproceedings{moore2015generalized,
  title={A generalized extended kalman filter implementation for the robot operating system},
  author={Moore, Thomas and Stouch, Daniel},
  booktitle={Intelligent Autonomous Systems 13: Proceedings of the 13th International Conference IAS-13},
  pages={335--348},
  year={2015},
  organization={Springer}
}

@article{SARTAYEVA2023103293,
title = {A survey on indoor positioning security and privacy},
journal = {Computers \& Security},
volume = {131},
pages = {103293},
year = {2023},
issn = {0167-4048},
doi = {https://doi.org/10.1016/j.cose.2023.103293},
url = {https://www.sciencedirect.com/science/article/pii/S0167404823002031},
author = {Yerkezhan Sartayeva and Henry C. B. Chan}
}

@ARTICLE{7321972,
  author={Li, Hong and He, Yunhua and Cheng, Xiuzhen and Zhu, Hongsong and Sun, Limin},
  journal={IEEE Communications Magazine}, 
  title={Security and privacy in localization for underwater sensor networks}, 
  year={2015},
  volume={53},
  number={11},
  pages={56-62},
  doi={10.1109/MCOM.2015.7321972}}

@ARTICLE{zhu2024protecting,
  author={Zhu, Yaping and Qiu, Ying and Wang, Junyuan and Hu, Jinming and Yan, Feng and Zhao, Shengjie},
  journal={IEEE Transactions on Network Science and Engineering}, 
  title={Protecting Position Privacy in Range-Based Crowdsourcing Cooperative Localization}, 
  year={2024},
  volume={11},
  number={1},
  pages={1136-1150},
  doi={10.1109/TNSE.2023.3321175}}

@article{shit2022privacy,
  title={Privacy-preserving cooperative localization in vehicular edge computing infrastructure},
  author={Chandra Shit, Rathin and Sharma, Suraj and Watters, Paul and Yelamarthi, Kumar and Pradhan, Biswajeet and Davison, Richard and Morgan, Graham and Puthal, Deepak},
  journal={Concurrency and Computation: Practice and Experience},
  volume={34},
  number={14},
  pages={e5827},
  year={2022},
  publisher={Wiley Online Library}
}

@article{le2025privacy,
  title={Privacy Preserving in Range-Based Cooperative Indoor Localization},
  author={Le, Yanfen and Wang, Shu and Lei, Ruolan and Yao, Heng},
  journal={IEEE Signal Processing Letters},
  year={2025},
  publisher={IEEE}
}

@article{gonzalez2020autonomous,
  title={Autonomous underwater vehicles: Localization, navigation, and communication for collaborative missions},
  author={Gonz{\'a}lez-Garc{\'\i}a, Josu{\'e} and G{\'o}mez-Espinosa, Alfonso and Cuan-Urquizo, Enrique and Garc{\'\i}a-Valdovinos, Luis Govinda and Salgado-Jim{\'e}nez, Tom{\'a}s and Escobedo Cabello, Jes{\'u}s Arturo},
  journal={Applied Sciences},
  volume={10},
  number={4},
  pages={1256},
  year={2020},
  publisher={MDPI}
}

@article{roumeliotis2002distributed,
  title={Distributed multirobot localization},
  author={Roumeliotis, Stergios I and Bekey, George A},
  journal={IEEE transactions on robotics and automation},
  volume={18},
  number={5},
  pages={781--795},
  year={2002},
  publisher={IEEE}
}

@article{lajoie2022towards,
  title={Towards collaborative simultaneous localization and mapping: a survey of the current research landscape},
  author={Lajoie, Pierre-Yves and Ramtoula, Benjamin and Wu, Fang and Beltrame, Giovanni},
  journal={Field Robotics},
  volume={2},
  pages={971--1000},
  year={2022},
  publisher={FRPS}
}

@article{fox2000probabilistic,
  title={A probabilistic approach to collaborative multi-robot localization},
  author={Fox, Dieter and Burgard, Wolfram and Kruppa, Hannes and Thrun, Sebastian},
  journal={Autonomous robots},
  volume={8},
  number={3},
  pages={325--344},
  year={2000},
  publisher={Springer}
}

@inproceedings{tang2025feasibility,
  title={On the Feasibility of Fingerprinting Collaborative Robot Network Traffic},
  author={Tang, Cheng and Barradas, Diogo and Hengartner, Urs and Hu, Yue},
  booktitle={International Conference on Availability, Reliability and Security},
  pages={95--117},
  year={2025},
  organization={Springer}
}

@inproceedings{denning2009spotlight,
  title={A spotlight on security and privacy risks with future household robots: attacks and lessons},
  author={Denning, Tamara and Matuszek, Cynthia and Koscher, Karl and Smith, Joshua R and Kohno, Tadayoshi},
  booktitle={Proceedings of the 11th international conference on Ubiquitous computing},
  pages={105--114},
  year={2009}
}

@article{oruma2023security,
  title={Security aspects of social robots in public spaces: a systematic mapping study},
  author={Oruma, Samson Ogheneovo and Ayele, Yonas Zewdu and Sechi, Fabien and R{\o}dsethol, Hanne},
  journal={Sensors},
  volume={23},
  number={19},
  pages={8056},
  year={2023},
  publisher={MDPI}
}

@InProceedings{pmlr-v242-tang24a,
  title={Uncertainty quantification of set-membership estimation in control and perception: Revisiting the minimum enclosing ellipsoid},
  author={Tang, Yukai and Lasserre, Jean-Bernard and Yang, Heng},
  booktitle={6th Annual Learning for Dynamics \& Control Conference},
  pages={286--298},
  year={2024},
  organization={PMLR}
}

@misc{angelopoulos2022gentleintroductionconformalprediction,
      title={A Gentle Introduction to Conformal Prediction and Distribution-Free Uncertainty Quantification}, 
      author={Anastasios N. Angelopoulos and Stephen Bates},
      year={2022},
      eprint={2107.07511},
      archivePrefix={arXiv},
      primaryClass={cs.LG},
      url={https://arxiv.org/abs/2107.07511}, 
}

@book{boyd2004convex,
  title={Convex optimization},
  author={Boyd, Stephen and Vandenberghe, Lieven},
  year={2004},
  publisher={Cambridge university press}
}

@article{boyd2011distributed,
  title={Distributed optimization and statistical learning via the alternating direction method of multipliers},
  author={Boyd, Stephen and Parikh, Neal and Chu, Eric and Peleato, Borja and Eckstein, Jonathan},
  journal={Foundations and Trends{\textregistered} in Machine learning},
  volume={3},
  number={1},
  pages={1--122},
  year={2011},
  publisher={Now Publishers, Inc.}
}

@Article{rs17152637,
AUTHOR = {Hari, Sai Krishna Kanth and Sundar, Kaarthik and Braga, José and Teixeira, João and Darbha, Swaroop and Sousa, João},
TITLE = {Robust Underwater Vehicle Pose Estimation via Convex Optimization Using Range-Only Remote Sensing Data},
JOURNAL = {Remote Sensing},
VOLUME = {17},
YEAR = {2025},
NUMBER = {15},
ARTICLE-NUMBER = {2637},
URL = {https://www.mdpi.com/2072-4292/17/15/2637},
ISSN = {2072-4292},
DOI = {10.3390/rs17152637}
}

\end{document}